\newif\ifdraft
\draftfalse

\documentclass[final,1p,times]{elsarticle}

\usepackage{url}
\usepackage{amssymb}
\usepackage{amsmath}

\usepackage{graphicx}
\usepackage{siunitx}
\usepackage{xcolor}
\usepackage{comment}

\usepackage{graphicx}
\usepackage{amsmath}
\usepackage{pifont}
\usepackage{amsmath, amssymb, amsfonts, amsthm}
\usepackage{graphicx}
\usepackage{amsfonts}       
\usepackage{nicefrac}       
\usepackage{microtype}      
\usepackage{mathtools}
\usepackage{comment}
\usepackage{pbox}
\usepackage{textcomp}
\usepackage[utf8]{inputenc}
\usepackage{multirow}
\usepackage{nameref}
\usepackage{textalpha}
\usepackage{booktabs} 
\usepackage{comment}
\usepackage{subfig}
\usepackage{threeparttable}
\usepackage{caption}

\usepackage{ragged2e} 

\usepackage[normalem]{ulem} 
\newcommand{\stkout}[1]{\ifmmode\text{\sout{\ensuremath{#1}}}\else\sout{#1}\fi}

\ifdraft
\newcommand{\added}[1]{\textcolor{blue}{#1}}
\newcommand{\deleted}[1]{\textcolor{red}{\sout{#1}}}
\newcommand{\replaced}[2]{\textcolor{blue}{#1} \textcolor{red}{\sout{#2}}}
\newcommand{\deletedfloat}[1]{}
\newcommand{\commented}[1]{\textcolor{blue}{#1}}
\else
\newcommand{\added}[1]{#1}
\newcommand{\deleted}[1]{}
\newcommand{\replaced}[2]{#1}
\newcommand{\deletedfloat}[1]{}
\newcommand{\commented}[1]{}
\fi

\journal{Computers in Biology and Medicine}

\begin{document}

\makeatletter
\def\ps@pprintTitle{%
     \let\@oddhead\@empty
     \let\@evenhead\@empty
     \def\@oddfoot
       {\hbox to \textwidth%
        {\ifnopreprintline\relax\else
        \@myfooterfont%
         \ifx\@elsarticlemyfooteralign\@elsarticlemyfooteraligncenter%
           \hfil Version accepted by \@journal \hfil%
         \else%
         \ifx\@elsarticlemyfooteralign\@elsarticlemyfooteralignleft%
           Version accepted by \@journal \hfill{}%
         \else%
         \ifx\@elsarticlemyfooteralign\@elsarticlemyfooteralignright%
           {}\hfill Version accepted by \@journal%
         \else%
           Version accepted by \@journal\hfill\@date%
         \fi%
         \fi%
         \fi%
         }%
       }%
     \let\@evenfoot\@oddfoot}
\makeatother

\begin{frontmatter}

\title{Enhancing clinical decision support with physiological waveforms -- a multimodal benchmark in emergency care}

\author[label1]{Juan Miguel Lopez Alcaraz}
\ead{juan.lopez.alcaraz@uol.de}

\author[label2]{Hjalmar Bouma}
\ead{h.r.bouma@umcg.nl}

\author[label1]{Nils Strodthoff\corref{cor1}}
\ead{nils.strodthoff@uol.de}
\cortext[cor1]{Corresponding author.}
\ead[url]{https://uol.de/en/ai4health}

\affiliation[label1]{organization={AI4Health Division, Carl von Ossietzky Universität Oldenburg}, 
            addressline={Ammerländer Heerstraße 114-118}, 
            city={Oldenburg}, 
            postcode={26129}, 
            state={Lower Saxony}, 
            country={Germany}}

\affiliation[label2]{organization={Department of Internal Medicine, Department of Acute Care, and Department of Clinical Pharmacy \& Pharmacology, University Medical Center Groningen}, 
            addressline={Hanzeplein 1}, 
            city={Groningen}, 
            postcode={9713}, 
            state={Groningen}, 
            country={Netherlands}}

\begin{abstract}
\deleted{AI-driven prediction algorithms have the potential to enhance emergency medicine by supporting
rapid and accurate decision-making regarding patient status and potential deterioration. We present
a dataset and benchmarking protocol to advance multimodal decision support in emergency care.
Using demographics, biometrics, vital signs, laboratory values, and electrocardiogram waveforms
as input, we train models to predict diagnoses and patient deterioration. Our diagnostic model
achieves area under the receiver operating curve (AUROC) scores above 0.8 for 609 out of 1428
conditions in a statistically significant manner, including cardiac (e.g., myocardial infarction) and
non-cardiac (e.g., renal disease, diabetes) diagnoses. The deterioration model scores AUROC
scores above 0.8 for 14 out of 15 targets, including critical events like cardiac arrest, mechanical
ventilation, ICU admission, and mortality. Notably, we provide a robust demonstration of the
positive impact of incorporating raw waveform data into decision support models. This study
introduces a unique dataset and baseline models that promote measurable progress in algorithmic
decision support for emergency care.}

\added{\textbf{Background:} AI-driven prediction algorithms have the potential to enhance emergency medicine by enabling rapid and accurate decision-making regarding patient status and potential deterioration. However, the integration of multimodal data, including raw waveform signals, remains underexplored in clinical decision support.}

\added{\textbf{Methods:} We present a dataset and benchmarking protocol designed to advance multimodal decision support in emergency care. Our models utilize demographics, biometrics, vital signs, laboratory values, and electrocardiogram (ECG) waveforms as inputs to predict both discharge diagnoses and patient deterioration.}

\added{\textbf{Results:} The diagnostic model achieves area under the receiver operating curve (AUROC) scores above 0.8 for 609 out of 1,428 conditions, covering both cardiac (e.g., myocardial infarction) and non-cardiac (e.g., renal disease, diabetes) diagnoses. The deterioration model attains AUROC scores above 0.8 for 14 out of 15 targets, accurately predicting critical events such as cardiac arrest, mechanical ventilation, ICU admission, and mortality.}

\added{\textbf{Conclusions:} Our study highlights the positive impact of incorporating raw waveform data into decision support models, improving predictive performance. By introducing a unique, publicly available dataset and baseline models, we provide a foundation for measurable progress in AI-driven decision support for emergency care.}

\end{abstract}

\begin{keyword}
Deep-learning \sep Emergency Department \sep  Machine learning \sep Medical Decision Support \sep Multimodal Data \sep Patient Diagnostics \sep Patient Deterioration.
\end{keyword}

\end{frontmatter}

\section{Introduction}

Artificial intelligence (AI) is an emerging field that has significantly enhanced healthcare and medicine \cite{Rajpurkar2022} in areas such as precision medicine \cite{johnson2021precision} and drug discovery \cite{dara2022machine}. The implementation of AI models is particularly relevant in acute and emergency care, where clinicians must address critical conditions within short time frames to make optimal clinical decisions. AI models can improve acute care by enabling early diagnosis \cite{hong2020prediction}, proposing tailored diagnostic workups \cite{harrou2020forecasting}, predicting admissions (including ward or intensive care unit (ICU)) \cite{covino2020predicting}, estimating survival rates \cite{wang2019machine}, and providing faster and low-cost diagnoses \cite{henna2022interpretable,alcaraz2024cardiolablaboratoryvaluesestimation}. There has been exponential growth in the number of scientific publications on new AI models. However, many of these studies have limitations, such as predicting only a limited set of specific conditions \cite{taylor2018predicting,hong2020prediction}, offering short time horizons for survival predictions, and requiring costly diagnostic tests \cite{hwang2019deep}. 

\added{Despite their potential, current state-of-the-art AI models for forecasting specific medical conditions rely primarily on standard clinical features (e.g., demographics, laboratory results) and neglect raw physiological signals such as waveforms. These signals are known to be highly informative for a diverse set of diseases \cite{strodthoff2024prospects} and represent one example of raw sensor data. They hold the promise of providing complementary information to standard clinical features to enhance the accuracy and robustness of prediction models.
The prediction models used vary widely, as noted in a recent meta-review \cite{Fleuren2020} in the context of sepsis prediction, but typically cannot be compared directly due to differences in experimental setup. This highlights the need for structured benchmarking protocols. Unlike mere evaluation, which often focuses on fixed metrics for a single study, benchmarking establishes common datasets, tasks, and performance baselines, ensuring that advancements in AI-driven clinical decision support are measurable, reproducible, and comparable across studies. Benchmarking, as one component of standardized performance evaluation, also plays a crucial role in validation and certification processes. These processes represent a significant hurdle that decision support systems must overcome to reach clinical practice.}

Healthcare datasets are essential for advancing medical research and innovation. However, much of this valuable data remains undisclosed due to commercial interests and privacy concerns. Despite these hurdles, there has been a recent surge in publicly accessible datasets aimed at facilitating the development and validation of machine learning models to tackle complex healthcare challenges \cite{strodthoff2024prospects,sun2024edcopilot,sundrani2023predicting,wornow2024ehrshot,hager2024evaluation}. Nonetheless, these datasets still have limitations, including size constraints \cite{sun2024edcopilot}, lack of open-source availability \cite{NEURIPS2023_8f61049e,lee2024multimodal}, and narrow scopes in terms of prediction tasks \cite{sun2024edcopilot,sundrani2023predicting,lee2024multimodal,wornow2024ehrshot,NEURIPS2023_8f61049e,hager2024evaluation}.

In this study, we investigate a multimodal decision support for emergency department (MDS-ED) using an open-source biomedical multimodal dataset which comprises various medical feature modalities collected within the first 1.5 hours of patient arrival at the ED department such as demographics, biometrics, vital parameters trends, laboratory trends, and electrocardiogram (ECG) waveforms, to predict a diverse and large range of clinical tasks beyond traditional predictive scenarios, such as cardiac and non-cardiac diagnoses, as well as patient deterioration such as mortalities, ICU admission, organ-specific failure, and mechanical ventilation to name a few. Compared to approaches using a single data modality, which may offer a limited view of the patient's condition, we demonstrate that fusing multiple data sources is critical for improving decision-making accuracy in the complex environment of emergency departments. This multimodal approach is essential for predicting immediate as well as long-term outcomes, enabling more comprehensive and accurate assessments of patient status.
\added{To summarize, we put forward the following technical contributions:
\begin{enumerate}
\item We propose the MDS-ED dataset along with a corresponding benchmarking protocol which stands out from literature approaches as detailed in Section~\ref{sec:rel}.
\item We put forward strong unimodular baseline models based on best-practices from the literature, see Section~\ref{sec:archs}.
\item We propose powerful multimodal prediction models that outperform these strong unimodal baselines in all the scenarios considered, as demonstrated in Section~\ref{sec:results}. To our knowledge, this is the first robust demonstration of the added value of incorporating raw waveforms into clinical prediction models in an acute care setting.
\end{enumerate}
}
\section{Methods}

\subsection{Clinical workflow and dataset creation} 

\begin{figure*}[!ht]
    \centering
    \includegraphics[width=\textwidth]{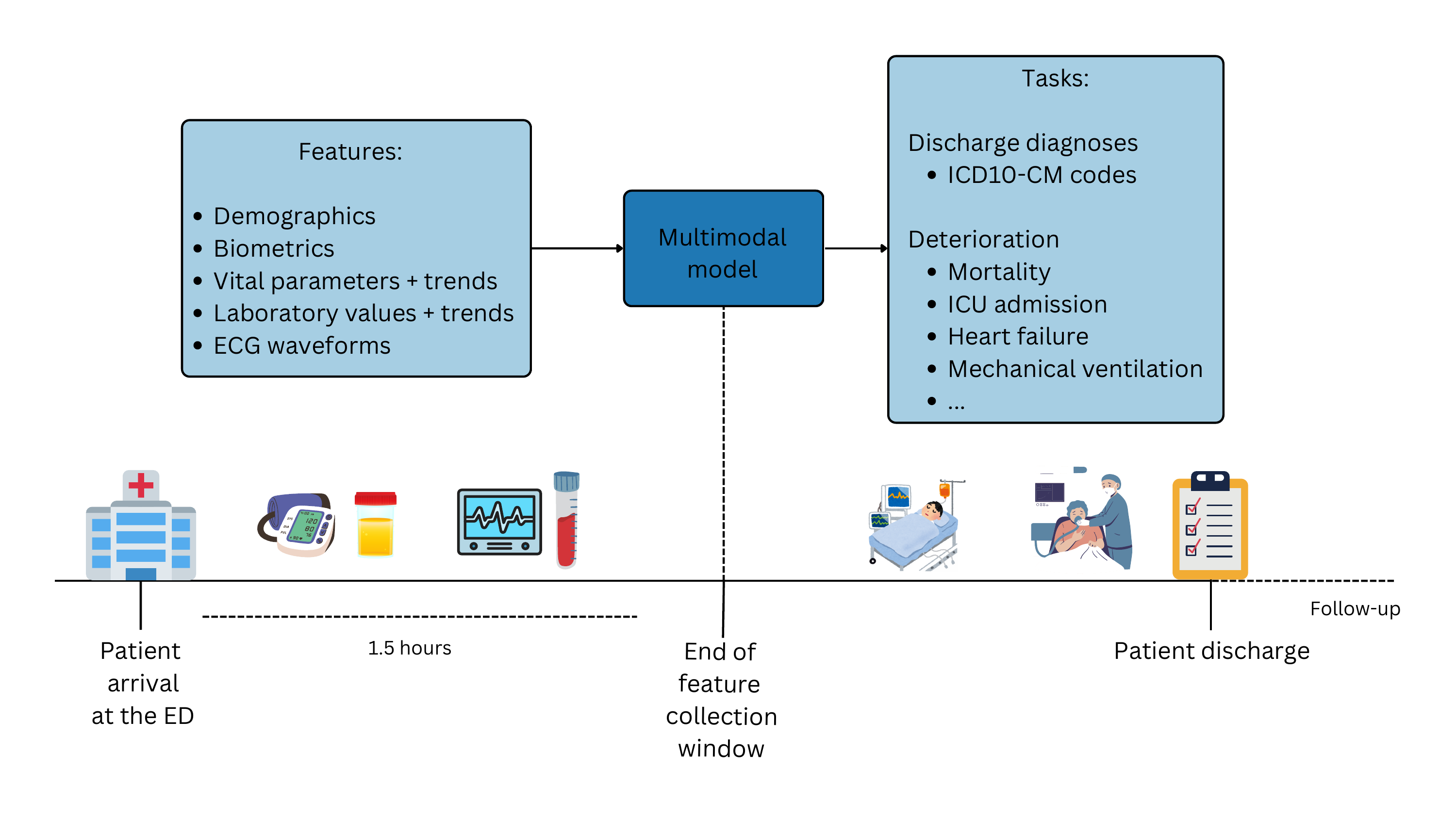}
    \caption{Pipeline outlines the MDS-ED \added{clinical} workflow, which involves feature collection encompassing patient demographics, biometrics (such as height, weight, and BMI), vital parameters and trends, laboratory values and trends, and ECG waveform data to address two clinically relevant prediction scenarios: predicting patient discharge diagnoses out of 1428 cardiac and non-cardiac ICD10-CM codes and predicting patient deterioration according to 15 clinical deterioration measures.}
    \label{fig:data_extraction_workflow}
\end{figure*}

Figure \ref{fig:data_extraction_workflow} presents a schematic illustration of the proposed MDS-ED pipeline which task focuses on a specific workflow with high clinical relevance, where for each ED visit, it encompasses a set of features collected from a window of 90 minutes from the patient's arrival at the ED to predict patient discharge diagnoses and deterioration through the stay. MDS-ED was created by linking ECG waveforms MIMIC database, specifically, MIMIC-IV-ECG \cite{MIMICIVECG2023} dataset to clinical features and outcomes as clinical ground truth from the MIMIC-IV and MIMIC-IV-ED dataset \cite{Johnson2023}. \added{The MIMIC dataset was chosen for its extensive collection of patient records across diverse demographic groups, encompassing thousands of features from clinical notes, lab results, vital signs, and full waveform data. This multimodal breadth enhances AI models predictive performance and clinical decision support.}

\begin{figure*}[!ht]
    \centering
    \includegraphics[width=\textwidth]{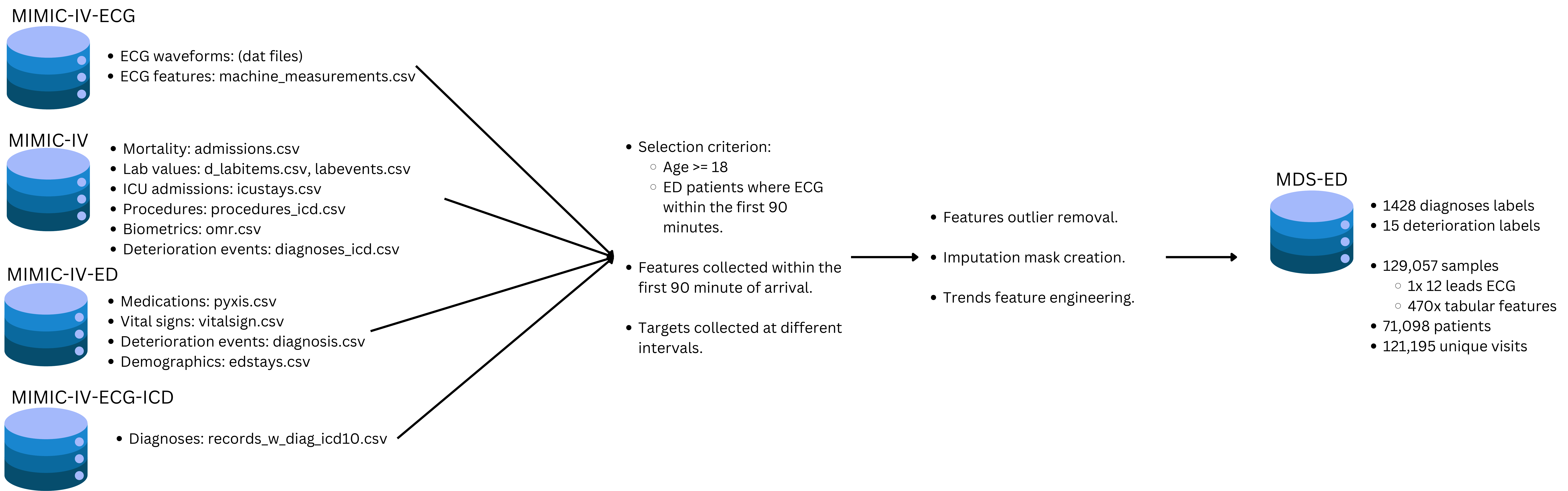}
    \caption{\added{Schematic representation summarizing the creation process of the MDS-ED dataset \cite{Lopez_Alcaraz_Strodthoff_2024} underlying this work. The process starts from four different source datasets (MIMIC-IV-ECG, MIMIC-IV, MIMIC-IV-ED, and MIMIC-IV-ECG-ICD) from which we select patients aged 18 years or older where an ECG was collected within the first 90 minutes of arrival at the ED department. Different target values collected at different intervals. On the resulting samples, we procedded to apply a features outlier removal which is primarily error-based, excluding unrealistic values, never-registered extremes, or negative values when the minimum is zero, see Appendix~\ref{app:dataset} and \cite{Lopez_Alcaraz_Strodthoff_2024} for details. Similarly, an imputation mask creation in which we apply median imputation from the train set to validation and test, adding binary masks to indicate imputed values, helping the model learn missingness patterns. Finally, we capture trends through engineered features such as summary statistics (mean, median, min, max, standard deviation), first and last values, rate of change, and slope of fitted linear model based on values within the first 90 minutes after arrival. The resulting MDS-ED dataset comprises 1428 diagnostic labels and 15 deteriorations labels across 129,057 samples from 71,098 patients collected from 121,195 unique visits. The input features cover a single 10s, 12-lead ECG in addition to 470 tabular features (excluding binary masking columns).}}
    \label{fig:dataset_creation}
\end{figure*}

\added{The MDS-ED dataset was constructed by integrating multiple sources to create a comprehensive resource for emergency medicine research. This process involved merging ECG recordings with diverse patient characteristics and clinical outcomes. Specifically, diagnostic ICD-10 codes and stratified splits were obtained from MIMIC-IV-ECG-ICD \cite{strodthoff2024prospects}, while ECG recordings were sourced from MIMIC-IV-ECG \cite{MIMICIVECG2023}. Clinical data were extracted from MIMIC-IV \cite{Johnson2023}, utilizing tables related to patient admissions, diagnoses, laboratory tests, ICU stays, and procedures. Additionally, emergency department-specific information, including stays, diagnoses, medication administration, vital signs, and triage details, was gathered from MIMIC-IV-ED \cite{Johnson2023}. The dataset creation process in visualized schematically in Figure \ref{fig:dataset_creation}. A complete description of the MDS-ED dataset along with the full preprocessing code for its creation is available on PhysioNet\cite{Lopez_Alcaraz_Strodthoff_2024}.}

\paragraph{Prediction tasks and targets}
In this work, we broadly categorize all prediction tasks into two groups: patient discharge diagnoses and patient deterioration. For discharge diagnoses, we follow MIMIC-IV-ECG-ICD \cite{strodthoff2024prospects}, which proposed a framework to predict discharge diagnoses based on a single electrocardiogram (ECG). To this end, we frame the task as a multilabel classification task in terms of international classification of diseases clinical modification 10 (ICD-10 CM) codes. As in \cite{strodthoff2024prospects}, we convert  ICD-9 to ICD-10 CM codes where necessary, convert codes to 5 digits, and propagate codes to parent codes up to the third digit. In this way, we obtain a total set of 1428 significantly populated ICD-10 CM codes covering a wide range of cardiac and non-cardiac conditions. For patient deterioration, we build on a consensus definition \cite{Mitchell2022} and investigate 15 deterioration events in a multilabel setting after the patient arrived at different intervals and after the 1.5-hour feature collection window: Six subtasks cover clinical deterioration within 24 hours and cover severe hypoxemia, extracorporeal membrane oxygenation (ECMO), use of vasopressors, use of inotropes, mechanical ventilation, and in-hospital cardiac arrest (IHCA). Two subtasks cover ICU admission within 24 hours and or within the entire stay, and finally, 7 subtasks cover mortality prediction at different horizons ranging from inpatient mortality, 24 hours, 7 days, 28 days, 90 days, 180 days, to 365 days.

\added{While ordinary predictive models often focus on immediate deterioration, our approach extends beyond standard scenarios by considering a broader temporal scope and a more comprehensive set of clinical events. Long-term prediction tasks, such as mortality at extended time horizons, pose unique challenges due to the evolving nature of patient conditions, while discharge diagnoses rely on accumulated clinical knowledge throughout the hospital stay and could be highly influential if identified early. By addressing both immediate and long-term risks, our approach ensures a more holistic and clinically relevant decision support framework. With the two main groups of diagnoses and deterioration, we aim to cover a wide spectrum of patient monitoring in the ED as well as long-term mortality across 1,443 unique target labels. See Table \ref{tab:tasks} in the supplementary material for a detailed discussion of the prediction targets.}

\paragraph{Features}
While aiming to include mainly data at triage, we also include irregularly sampled vital parameters and lab values captured during the 1.5h data collection window. To capture trends, we compute simple sample-wise statistical aggregation functions per vital parameter and laboratory values, namely mean, median, minimum, maximum, standard deviation, first and last values, rate of change of them between first and last value, as well as the slope of a linear model fitted on the minutes difference between value collection and arrival as independent variable and the actual values as dependent variables. We follow \cite{moor2023predicting} to select an appropriate set of laboratory values that avoids a bias towards particular prediction targets. We obtain a total of 470 features of clinical routine data across all feature modalities in a tabular format plus a 12-lead 10-second ECG waveform per sample. Similarly, the ECG feature set is composed of RR interval, P-onset, QRS-onset, QRS-end, and T-end in milliseconds (ms), whereas P-axis, QRS-axis, and T-axis in degrees.

\paragraph{\added{Train-test splits}}
MDS-ED consists of data from 71,098 patients, encompassing 121,195 unique visits, and a total of 129,057 samples. We leverage the stratified splits provided along with the dataset accompanying \cite{strodthoff2024prospects}, which includes stratification based on gender, age bins, and discharge diagnoses. We distribute the total number of 20 stratified folds into train, validation, and test following ratios of 18:1:1.

\subsection{Related work} 
\label{sec:rel}

\begin{table*}[bt!]
\caption{Direct comparison with related works in terms of dataset size, features, availability, and number of target labels.}
\label{tab:comparison}
\centering
\scriptsize
\begin{tabular}{l l c c c c c c}
\toprule
\textbf{Group} & \textbf{Detail} & \textbf{MIMIC-ED-Assist} & \textbf{VitalML} & \textbf{MEME} & \textbf{EHRSHOT} & \textbf{MC-BEC} & \textbf{MDS-ED} \\ 
\midrule
Source & & \cite{sun2024edcopilot} & \cite{sundrani2023predicting} & \cite{lee2024multimodal} & \cite{wornow2024ehrshot} & \cite{NEURIPS2023_8f61049e} & This work \\
\midrule
Population & & ED & ED & Longitudinal & Longitudinal & ED & ED \\
\midrule
\multirow{2}{*}{Size} & Patients & 25714 & N/A & 947028 & 6739 & 63389 & 71098 \\
& Visits & 32356 & 19847 & N/A & 921499 & 102731 & 121195 \\
\midrule
\multirow{7}{*}{Features} & Demographics & \ding{51} & \ding{51} & \ding{51} & \ding{51} & \ding{51} & \ding{51} \\
& Biometrics & \ding{55} & \ding{55} & \ding{55} & \ding{55} & \ding{55} & \ding{51} \\
& Vital parameters & \ding{51} & \ding{51}($\mathbb{T}$) & \ding{51}($\mathbb{T}$) & \ding{55} & \ding{51}($\mathbb{T}$) & \ding{51}($\mathbb{T}$) \\
& Lab. values & \ding{51} & \ding{55} & \ding{55} & \ding{51} & \ding{51}($\mathbb{T}$) & \ding{51}($\mathbb{T}$) \\
& Waveforms & \ding{55} & \ding{51}($\mathbb{E}$) & \ding{55} & \ding{55} & \ding{51} & \ding{51} \\
& Chief complaint & \ding{51} & \ding{55} & \ding{51} & \ding{55} & \ding{51} & \ding{55} \\
& Medications & \ding{55} & \ding{55} & \ding{51} & \ding{51} & \ding{51} & \ding{55} \\
\midrule
Tasks & Labels & 3 & 3 & 4 & 15 & 7 & 1443 \\
\midrule
Availability & Open source & \ding{51} & \ding{51} & \ding{55} & \ding{51} & \ding{55} & \ding{51} \\
\bottomrule
\end{tabular}
\begin{tablenotes}
\item We use diverse symbology to express the contribution where \ding{51} = available, \ding{55} = unavailable, $\mathbb{E}$ = available in the form of embeddings, and $\mathbb{T}$ = available in the form of trends or at least two sampled values.
\end{tablenotes}
\end{table*}

Table \ref{tab:comparison} contains a direct comparison with related works in terms of population, dataset size, features, availability, and number of target labels. \cite{sun2024edcopilot} proposed the MIMIC-ED-Assist benchmark to reduce ED length of stay by flagging high-risk patients using triage features. \cite{sundrani2023predicting} proposed VitalML to predict deterioration in the next 90 minutes from the first 15 minutes of monitoring. \cite{lee2024multimodal} proposed a multiple embedding model for \replaced{EHR data}{electronic health records (EHR)} (MEME) to generate pseudo-notes from tabular data and predict ED disposition, discharge location, ICU admission, and inpatient mortality utilizing a natural language processing approach. \cite{wornow2024ehrshot} proposed a few shot foundational model EHRSHOT approach to predict operational outcomes, developed diagnoses, as well as anticipate lab test results and image findings. \cite{NEURIPS2023_8f61049e} proposed MC-BEC to predict patient deterioration, disposition, and revisit from features collected including diagnoses from the first 15 minutes after the patient is assigned to a room.  

Overall, our proposed MDS-ED main contributions in terms of the dataset are: \textbf{1) Comprehensive size:} MDS-ED is situated in the first position regarding a large number of patients and second in the number of visits in the open-source domain, despite restricting towards the investigated setting of the first 1.5 hours of ED arrival. \textbf{2) Diverse input features:} MDS-ED is situated in the first position in terms of feature modalities in the open-source domain as it incorporates a comprehensive set of features, including demographics, biometrics, vital parameter trends, laboratory values trends, and ECG waveforms. This is more extensive than most compared datasets. Intentionally, we decided to exclude chief complaints as their unstructured nature presents a more challenging step towards external validation with the addition of more variance, see \cite{lee2024multimodal}. Similarly, we exclude previous patient medications as they involve irregularities in administration times, and dosages, as well as potential bias towards task prediction, e.g., certain medications are given for specific gravity and stages of diagnoses. \textbf{3) Comprehensive prediction targets:} MDS-ED proposes 1,443 target labels, significantly more than other datasets, which typically cover fewer and too narrow in scope tasks. \textbf{4) Accessibility:} Similar to \cite{sun2024edcopilot,sundrani2023predicting,wornow2024ehrshot}, MDS-ED is open source, encouraging further research and collaboration.

\subsection{Model architectures} 
\label{sec:archs}

To propose an initial benchmark, we experiment with diverse settings based on data modalities input such as (1) clinical routine data, (2) ECG features only, (3) ECG waveforms only, (4) ECG features + clinical routine data, and (4) ECG waveforms + clinical routine data. For waveforms (3) and (5), inspired by recent successful applications in the field of physiological time series \cite{strodthoff2024prospects,wang2023s4sleep,mehari2023towards} we employed an S4 classifier based on structured state space models with four layers \cite{Gu2021EfficientlyML} (\added{relying on hyperparameter choices from prior work\cite{mehari2023towards}, i.e, } \replaced{using}{with} state dimension 8 and model dimension 512)\added{, which has been shown to outperform more conventional deep learning architectures, such as modern convolutional neural networks, recurrent neural networks or transformer models, as signal encoder \cite{strodthoff2024prospects,wang2023s4sleep,mehari2023towards}}. For the multimodal models in (5), we use a 3-layer MLP as tabular encoder, where the three categorical features (gender, race, acuity score) are first processed through appropriate embedding layers. The outputs of the signal encoder and the tabular encoder are then fused through concatenation. \replaced{We}{For deep learning models with clinical routine data, we} apply median imputation \replaced{using training set statistics. In addition, we introduce a binary column per input variable to indicate imputed values. These binary columns are provided as additional inputs next to the original tabular features.}{learned from train set to validation and test for each column and we apply a masking strategy in which we add an additional binary column to indicate which values were imputed to the model.} 
\added{In \ref{app:mask_ablation} in the supplementary material, we present an ablation study comparing different imputation strategies.} \added{A systematic exploration of other tabular encoder models or more complex multimodal fusion schemes beyond the basic choices from above is deferred to future work}. For (1), (2) and (4) we fit extreme gradient boosting (XGBoost) decision trees\added{, which widely accepted state-of-the-art for prediction models on tabular data}. \added{We used standard hyperparameter defaults commonly applied in gradient-boosting methods.} Additional details on model architectures and model hyperparameters can be found in \ref{app:models} Table \ref{tab:s4-hyperparameters}.

\subsection{Training and evaluation} 

For the S4 models, we used \replaced{AdamW \cite{loshchilov2018decoupled}}{Adam Weight Decay Optimization (AdamW)} as the optimizer, with both learning rate and weight decay set to 0.001, and maintained a constant learning rate schedule. The training was conducted with a batch size of 64 samples over 20 epochs \deleted{which usually converged earlier and model selection on the validation set.} \added{The hyperparameter settings coincide with those from an earlier unimodal investigation \cite{strodthoff2024prospects}, which underscores the robustness of our findings without excessive hyperparameter tuning. Overfitting was mitigated through model selection based on validation performance, making the total number of epochs irrelevant as long as it was sufficiently large for convergence.}  The training objective was to minimize binary cross-entropy loss. For all the models, we evaluate performances on the macro average across all areas under the respective receiver operating curves (AUROC) (macro AUROC). To assess statistical uncertainty resulting from the test set's finite size and specific composition, we employ empirical bootstrap on the test set with $n=1000$ iterations. We report 95\% confidence intervals for both macro AUROC and individual label AUROCs. Given \cite{strodthoff2024prospects} evaluation protocol, we use only the first record per visit per patient to avoid bias. \added{We use macro AUROC as the primary metric due to its status as most widely used ranking-based metric. It characterize a model's overall discriminative power without the necessity to fix custom decision thresholds. The suitability of AUROC compared to the other commonly used metric in the presence of label imbalance, the area under the precision recall curve (AUPRC), was confirmed in a recent theoretical and empirical work \cite{mcdermott2024closer}.} Additional details of training and evaluation procedures can be found in \ref{app:models}.

\section{Results}
\label{sec:results}

\subsection{Benchmarking predictive performance}\label{subsec: benchmarking pp}

\begin{table*}[bt!]
\caption{Predictive performance (macro AUROC with 95\% bootstrap confidence intervals) across all scenarios for unimodal models as well as multimodal prediction models. Where the clinical routine data encompasses a comprehensive set of 470 variables from diverse feature modalities such as demographics, biometrics, vital parameters trends, and laboratory values trends. All the scenarios can also be split into model architecture, either tree-based or deep-learning models.}
\label{tab:predictive performance}
\centering
\scriptsize
\begin{tabular}{ccc}
\toprule
& \textbf{Diagnoses} & \textbf{Deterioration} \\
\midrule
& \multicolumn{2}{c}{\textbf{Unimodal models}} \\ 
\midrule
Clinical routine data (tree-based) & 0.7473 (0.7459, 0.7484) & 0.8365 (0.8179, 0.8546) \\ 
\midrule
ECG features only (tree-based) &  0.6366 (0.6352, 0.6374) &  0.7385 (0.7114, 0.7588) \\ 
\midrule
ECG waveforms only (deep-learning) & 0.7678 (0.7636, 0.7718) & 0.8565 (0.8410 , 0.8707) \\
\midrule
& \multicolumn{2}{c}{\textbf{Multimodal models}} \\
\midrule
ECG features + clinical routine data (tree-based) & 0.7685 (0.7676, 0.7705) & 0.8656 (0.8498, 0.8830) \\ 
\midrule
\textbf{ECG waveforms + clinical routine data (deep-learning)} & \textbf{0.8256 (0.8222,0.8288)} & \textbf{0.9115 (0.8991,0.9222)} \\ 
\bottomrule
\end{tabular}
\end{table*}

Table \ref{tab:predictive performance} presents the predictive performance of both unimodal and multimodal models across the diagnoses and deterioration settings. For diagnoses prediction, the multimodal model that integrates ECG waveforms and clinical routine data achieved a macro average across all areas under the respective receiver operating curves (AUROC) (macro-AUROC) of 0.8256 (0.8222,0.8288), which is notably higher than the rest of the models. Similarly, in the deterioration prediction task, the multimodal model that integrates ECG waveforms and clinical routine data achieved an AUROC of 0.9115 (0.8991,0.9222), which also overcomes the rest of the models. In the following discussion, we restrict ourselves to the discussion of the most comprehensive multimodal model based on ECG waveforms and clinical routine data. \added{In  \ref{app:advantages_multimodal}, we present a more finegrained comparison across different ICD10 chapters and specific deterioration targets, where the multimodal deep learning consistently outperforms unimodal baseline models.}

\subsection{Task-dependent predictive performance}\label{subsec: task-dependent pp}

\begin{table*}[ht!]
\centering
\tiny

\caption{Best-performing individual statements organized according to selected ICD chapters underscoring the breadth of accurately predictable statements. The table shows the four best-performing individual statements per ICD chapter (8 for chapter IX (Circulatory system diseases), 4 for the rest), where we show only AUROC scores above 0.85 and where also the lower bound of the 95\% bootstrap confidence interval exceeds 0.80. To showcase the breadth of reliably predictable statements, we list only the best-performing statement per 3-digit ICD code.}
\label{tab:big_all}

\begin{tabular}{p{6.2cm}p{6.2cm}}
\toprule
\textbf{Code: AUROC [instances]. Description} & \textbf{Code: AUROC [instances]. Description} \\  
\hline
\midrule

\multicolumn{2}{c}{\textbf{IX: Circulatory system.  0.8761 (126/181 $>$0.80 lower bound)}} \\ 
\midrule
I071: 0.9461 [287] Rheumatic tricuspid insufficiency & I080: 0.8522 [770] Rheumatic mitral and aortic valve disorders \\
I110: 0.9044 [3185] Hypertensive heart disease with heart failure & I120: 0.9756 [3349] Hypert. CKD (stage 5 or with end-stage renal disease) \\
I130: 0.9426 [2994] Hypertensive heart and kidney disease with heart failure and CKD & I160: 0.9524 [427] Hypertensive urgency \\
I200: 0.8934 [810] Unstable angina & I2109: 0.9734 [401] STEMI anterior wall coronary artery \\
\midrule

\multicolumn{2}{c}{\textbf{II: Neoplasms. 0.8572 (34/53 $>$0.80)}} \\ 
\midrule
C228: 0.9628 [235] Malignant neoplasm of liver, primary & C259: 0.8991 [275] Malignant neoplasm of pancreas \\
C341: 0.8739 [463] Malignant neoplasm of upper lobe or lung & C5091: 0.8544 [425] Malignant neoplasm of breast, female \\
\midrule

\multicolumn{2}{c}{\textbf{XIX: Injury, poisoning, and certain other consequences of external causes. 0.8479 (37/86 $>$0.80)}} \\ 
\midrule
S0100: 0.9161 [258] Unspecified open wound of scalp & S065: 0.8767 [783] Traumatic subdural hemorrhage \\ 
S12: 0.9356 [392] Fracture of cervical vertebra and other parts of neck & S2249: 0.9198 [409] Multiple fractures of ribs \\
\midrule

\multicolumn{2}{c}{\textbf{I: Certain infectious and parasitic diseases. 0.8346 (19/52 $>$0.80)}} \\ 
\midrule
A0472: 0.9374 [235] Enterocolitis due to Clostridium difficile & A084: 0.8709 [352] Viral intestinal infection, \\ 
A40: 0.9347 [339] Streptococcal sepsis & A4151: 0.9522 [626] Sepsis due to Escherichia coli [E. coli] \\
\midrule

\multicolumn{2}{c}{\textbf{IV: Endocrine, nutritional, and metabolic diseases. 0.8336 (52/107 $>$0.80)}} \\ 
\midrule
E1022: 0.9661 [309] Type 1 diabetes mellitus with diabetic CKD & E1121: 0.9466 [707] Type 2 diabetes mellitus with diabetic nephropathy \\ 
E1342: 0.9271 [1809] Diabetes mellitus with diabetic polyneuropathy & E43: 0.9329 [1417] Severe protein-calorie malnutrition \\
\midrule

\multicolumn{2}{c}{\textbf{III: Diseases of the blood and blood-forming organs. 0.8335 (20/45 $>$0.80)}} \\ 
\midrule
D500: 0.8557 [897] Cronic Iron deficiency anemia   & D6181: 0.8761 [1503] Pancytopenia \\ 
D62: 0.8843 [4514] Acute posthemorrhagic anemia & D630: 0.8920 [589] Anemia in neoplastic disease \\
\midrule

\multicolumn{2}{c}{\textbf{XI: Diseases of the digestive system. 0.827 (40/119 $>$0.80)}} \\ 
\midrule
K264: 0.9189 [255] Chronic duodenal ulcer with hemorrhage & K3181: 0.9384 [419] Angiodysplasia of stomach and duodenum \\ 
K521: 0.9290 [257] Toxic gastroenteritis and colitis & K566: 0.8958 [598] Other and unspecified intestinal obstruction \\
\midrule

\multicolumn{2}{c}{\textbf{XIV: Diseases of the genitourinary system. 0.8253 (18/39 $>$0.80)}} \\ 
\midrule
N08: 0.9393 [798] Glomerular disorders & N12: 0.8941 [325] Tubulo-interstitial nephritis, not acute or chronic \\ 
N170: 0.9092 [1773] Acute kidney failure with tubular necrosis & N183: 0.8625 [4734] Chronic kidney disease, stage 3 \\
\midrule

\multicolumn{2}{c}{\textbf{X: Diseases of the respiratory system. 0.8226 (33/71 $>$0.80)}} \\ 
\midrule
J1521: 0.9202 [275] Pneumonia due to staphylococcus aureus & J39: 0.9144 [276] Other diseases of upper respiratory tract \\ 
J440: 0.8936 [440] Chronic obstructive pulmonary disease with acute lower respiratory infection & J690: 0.8968 [2231] Pneumonitis due to inhalation of food and vomit \\
\midrule

\multicolumn{2}{c}{\textbf{V: Mental, behavioral and neurodevelopmental disorders. 0.8111 (32/71 $>$0.80)}} \\ 
\midrule
F0280: 0.8556 [1540] Dementia  & F0390: 0.8565 [3690] Unspecified dementia \\ 
F1012: 0.9591 [625] Alcohol abuse with intoxication & F1110: 0.8735 [666] Opioid abuse, uncomplicated \\
\midrule

\multicolumn{2}{c}{\textbf{XII: Diseases of the skin and subcutaneous tissue. 0.8087 (10/25 $>$0.80)}} \\ 
\midrule
L0311: 0.8646 [573] Cellulitis of other parts of limb & L408: 0.8933 [292] Other psoriasis \\ 
L8915: 0.8920 [1209] Pressure ulcer of sacral region & L974: 0.9093 [336] Non-pressure chronic ulcer of heel and midfoot \\
\midrule

\multicolumn{2}{c}{\textbf{VI: Diseases of the nervous system. 0.7955 (19/67 $>$0.80)}} \\ 
\midrule
G309: 0.8508 [1192] Alzheimer's disease & G4030: 0.9374 [244] Idiopathic epilepsy and epileptic syndromes \\
G609: 0.8772 [816] Hereditary and idiopathic neuropathy & G8190: 0.9251 [265] Hemiplegia \\
\midrule

\multicolumn{2}{c}{\textbf{XVIII: Symptoms, signs and abnormal clinical and laboratory findings 0.7813 (35/171 $>$0.80)}} \\ 
\midrule
R000: 0.8507 [2714] Tachycardia, unspecified & R042: 0.8745 [316] Hemoptysis \\ 
R0600: 0.8502 [1999] Dyspnea & R1032: 0.8956 [453] Left lower quadrant pain \\
\midrule

\multicolumn{2}{c}{\textbf{XIII: Diseases of the musculoskeletal system and connective tissue. 0.7405 (14/68 $>$0.80)}} \\ 
\midrule
M100: 0.9106 [246] Idiopathic gout & M129: 0.8933 [495] Arthropathy \\ 
M1A9: 0.9000 [253] Chronic gout, unspecified & M810: 0.8439 [4398] Age-related osteoporosis without fracture \\

\bottomrule
\end{tabular}
\end{table*}

Table \ref{tab:big_all} demonstrates the diagnoses that can be predicted most accurately by the model, which we organized based on ICD chapters. The algorithm's predictive performance ranges from an AUROC of 0.7405 for the musculoskeletal system and connective tissue (XIII) chapter to 0.8761 for the circulatory system (IX) chapter. In total, the model predicted 609 out of 1,428 individual ICD diagnoses with high accuracy, which we define in this work as conditions where the lower bound of the 95\% bootstrap confidence interval for AUROC exceeds 0.80. 
\added{A more fine-grained comparison of model performance across different input settings in the deterioration prediction task reveals notable trends in Table \ref{tab:multimodalvs1}, where the inclusion of ECG waveforms alongside clinical routine data consistently improves performance over ECG features and clinical routine data, with relative improvements ranging from 2.36\% to 13.06\%. The highest gains are observed for XII (Skin, 13.06\%), VI (Nervous, 10.75\%), and XIX (Injury, Poisoning, 9.95\%), suggesting that continuous waveform signals provide critical insights over features into these non-cardiac groups. In contrast, smaller improvements are seen for III (Blood, 2.36\%) and IV (Endocrine, 5.25\%), indicating that static ECG features might already capture most of the predictive signal in these cases.}

\begin{table}[ht!]
\centering
\scriptsize
\caption{Detailed breakdown of model performance in the deterioration category.}
\label{tab:big_all_deterioration}
\begin{tabular}{ll}
\toprule
\textbf{Label: AUROC [instances]} & \textbf{Label: AUROC [instances]} \\  
\hline
\midrule
\multicolumn{2}{c}{\textbf{Clinical deterioration. 0.9070 (5/6 $>$ 0.80) lower bound}} \\ 
\midrule
Severe hypoxemia: 0.6980 [555] & ECMO: 0.9355 [166] \\
Vasopressors: 0.9239 [1171] & Inotropes: 0.9400 [375] \\
Mechanical ventilation: 0.9590 [4223] & IHCA: 0.9859 [623] \\
\midrule
\multicolumn{2}{c}{\textbf{ICU admission. 0.9063 (2/2 $>$ 0.80)}} \\
\midrule
ICU 24 hours: 0.9147 [15677] & ICU overall: 0.8979 [18860] \\
\midrule
\multicolumn{2}{c}{\textbf{Mortality. 0.9168 (7/7 $>$ 0.80)}} \\
\midrule
In-hospital: 0.9423 [99] & 24 hours: 0.9600 [816] \\
7 days: 0.9429 [2273] & 28 days: 0.9115 [5150] \\
90 days: 0.8952 [9297] & 180 days: 0.8894 [12839] \\
365 days: 0.8768 [17756] & \\
\bottomrule
\end{tabular}
\end{table}

Table \ref{tab:big_all_deterioration} showcases the deterioration model's predictive performance across clinical deterioration, ICU admission, as well as both short and long-term mortality with many AUROC scores significantly exceeding the 0.80 threshold. Overall, for clinical deterioration, the model achieves an AUROC of 0.9070, whereas, for ICU admissions, the model reports an overall AUROC of 0.9063, and for mortality predictions also exhibits high performance, with an overall AUROC of 0.9168.
\added{In Tables \ref{tab:multimodalvs1} and \ref{tab:multimodalvs2} in the supplementary material, we present a  more fine-grained comparison of model performance across different input settings for the deterioration prediction task. As in the diagnostic case, the multimodal model leveraging waveforms outperforms the corresponding model leveraging only ECG features in addition to clinical routine data in all cases, most notably in the clinical deterioration category with a relative improvement of more than 9\%. On the most finegrained level in Table \ref{tab:multimodalvs2}, one observes relative improvements ranging from 1.41\% (365-day mortality) to 38.08\% (severe hypoxemia). The most substantial gains occur for severe hypoxemia (38.08\%) and inotrope administration (16.57\%), suggesting that waveform-based inputs contribute significantly to an early identification of respiratory failure and cardiac dysfunction. Conversely, for longer-term mortality outcomes (e.g., 90-day to 365-day), the added benefit of waveform data diminishes, as patient trajectories become increasingly influenced by non-acute factors that are already reliably captured by clinical routine features and ECG features.}

\section{Discussion} 

\subsection{Impact of data modalities}
From the results presented in the previous section, we can draw several conclusions: Firstly, the results demonstrate that multimodal models, which integrate diverse data types, offer superior performance in both diagnostic and deterioration tasks (row 4\&5 vs. the rest). Secondly, in the diagnoses task as well as in the deterioration task, the use of ECG raw waveforms instead of ECG features improves the performance in a statistically significant manner (row 4 vs. row 5), finding which is not in line with \cite{sundrani2023predicting}. To the best of our knowledge, this is the first statistically robust demonstration of the added value of raw ECG waveform input against ECG features for clinically relevant prediction tasks such as diagnoses and deterioration prediction. Thirdly, for the unimodal models, in the deterioration task, the clinical routine data model outperforms ECG features only and ECG waveforms only, however, in the diagnoses task, the ECG waveforms only outperforms the other \replaced{two}{2} settings, we hypothesize that for the deterioration task, the clinical routine data apart of including a rich set of clinical features (demographics, biometrics, vital parameters trends, and laboratory values trends) against only an single ECG either in features or waveform, it also includes trends over time which aligns with the task definition of deterioration. Despite this, we believe that a single ECG snapshot can achieve high performances for both tasks, but also we believe that the inclusion of multiple ECGs over time instead of just a single snapshot would allow us to capture more meaningful deterioration and potentially diagnoses trends.
\added{Fourthly, direct comparison to literature results is challenging due to the lack of publicly available datasets and the corresponding scarcity of models that integrate both clinical routine data and waveform data. Among existing studies, \cite{NEURIPS2023_8f61049e} is one of the few incorporating waveform-based models, but differences in dataset construction and task formulation limit direct comparisons. Consequently, from a methodological perspective, we evaluate our approach using state-of-the-art models, including an ensemble tree-based model for tabular data and a signal encoder with proven efficacy, as previously presented in \cite{strodthoff2024prospects}.}

\subsection{Clinical significance}

From a clinical point of view, we conclude that our diagnostic models exhibit high predictive capabilities for acute, chronic, as well as management conditions. Acute conditions like ST elevation myocardial infarction, acute pericarditis, persistent atrial fibrillation, and acute kidney failure require rapid diagnosis and immediate intervention to prevent severe outcomes. The model's strong performance in identifying these conditions highlights its potential to significantly impact patient care through timely and accurate diagnosis. Conversely, chronic conditions such as hypertensive heart and chronic kidney disease, ischemic cardiomyopathy, dilated cardiomyopathy, and end-stage renal disease benefit from ongoing management and long-term treatment strategies. The model's ability to effectively predict these chronic conditions underscores its utility in supporting sustained patient monitoring and management, ultimately improving patient outcomes through consistent and reliable diagnostic support. Finally, detecting management conditions like thrombosis of prosthetic devices enhances ongoing care and ensures timely intervention. As demonstrated in \cite{strodthoff2024prospects}, even the model based only on ECG waveforms can reliably predict cardiac but most notably non-cardiac conditions. This applies even more to the multimodal modal that involves clinical features as well as ECG waveforms.

Our deterioration model has major clinical relevance by enhancing early recognition of patient deterioration and supporting physician decisions in the emergency department. Its high predictive performance across various scenarios effectively identifies patients at risk for severe complications, which is crucial in acute and emergency medicine. The model's ability to reliably predict clinically relevant indicators, such as vasopressor use, inotrope use, mechanical ventilation, and in-hospital cardiac arrest, enables healthcare providers to anticipate these events and implement early, tailored interventions. Additionally, its robust performance in predicting ICU admissions allows for better resource allocation, ensuring that patients who need intensive care receive it promptly while preventing unnecessary admissions. The model accurately predicts 24-hour in-hospital mortality and accurately predicts 365-day mortality, supporting clinical decision-making in prioritizing high-risk patients and discussing prognosis with families. Long-term mortality predictions aid in advanced care planning, such as deciding against ICU admission for patients with a short expected lifespan after discharge, and inform post-discharge planning and follow-up care to prevent readmissions and improve long-term health outcomes.

\subsection{Limitations}
Despite the promising results, there are several limitations to consider. 
\added{Firstly, diagnostic labels were inferred from ICD10 codes, which are subject to biases and incompleteness as they are primarily used for billing purposes and are not assigned purely based on clinical considerations. Clinical deterioration labels are comparably reliable as they are sourced from the same hospital system. Post-discharge mortality might again be subject to inaccuracies as it originates from a different source system.}

\added{Secondly, the presented models rely on the specifics of Beth Israel Deaconess Medical Center in terms of patient demographics, available resources, process characteristics, and data collection practices. The presented results allow statements about the generalizability of the proposed models to unseen patients with well-defined statistical uncertainty measures, as far as this is possible. However, it does not allow any statements about the transferability of these findings to other hospitals. It is important to note that generalization to other sites is challenging not only due to different patient demographics but also due to available resources, process characteristics, and data collection practices. This extends as far as certain labels, such as ICU admission, that are directly affected by process characteristics. The challenges of external validation in the context of complex clinical decision support systems are nicely summarized in a recent article \cite{laRoiTeeuw2024}.}

\replaced{Thirdly, all considered prediction scenarios rely on }{Secondly, the reliance on} data collected within the first 1.5 hours of the ED visits\added{, which} may limit the model’s applicability. We deliberately decided to focus on this setting to keep the dataset homogeneous. \added{There are two directions that are worth exploring in future work. On the one hand,}\deleted{However,} there is \deleted{also} a clear clinical case for a prediction model only incorporating information available during triage. \added{On the other hand, it might be instructive to extend the time horizon beyond 1.5 hours in particular to improve the performance for deterioration prediction tasks.} \replaced{In both cases, datasets similar}{A similar dataset} to the one proposed could be created, but this task is left for future research. 

\added{Finally, structured benchmarks through open-source datasets is only a first step towards measurable progress in the domain. Translating these research findings into clinical applications poses significant additional challenges. For example, deploying large-scale models will require substantial investments in hardware, in particular for real-time applications. Clinical decision support systems have to be carefully validated on diverse patient population and have to undergo certification, have to be properly integrated into clinical workflows and have to be closely monitored after deployment to detect early signs of performance deterioration due to time-dependent dataset shifts.}

\added{To ensure clinical utility, future work should prioritize validation across diverse patient populations and healthcare settings. While many AI algorithms are developed and published, only a small fraction undergo prospective validation or certification for real-world deployment. Regulatory approval and integration into existing clinical workflows present significant challenges, which may impact clinician trust and adoption. A structured benchmarking process could help to streamline this by ensuring a more standardized performance evaluation.}

\subsection{Future research directions}

The proposed dataset and benchmark protocol are based on input parameters that are typically available in clinical workflows and address a set of comprehensive and clinically meaningful prediction tasks. We therefore envision it to represent a meaningful baseline to benchmark prediction algorithms in the field. \added{In addition to addressing the limitations mentioned in the previous sections, we consider pursuing the following technical research directions as particularly promising:}

\added{Future research directions could focus on expanding the scope of input variables beyond the current feature set, incorporating additional clinical and physiological signals to enhance predictive performance.} An obvious extension \added{along these lines} would be to include free-text data \replaced{as in the recent of Hager et al}{along the lines of} \cite{hager2024evaluation}\added{. This could enhance the predictive performance by capturing richer clinical context, uncovering implicit relationships, and incorporating nuanced information that structured data alone might miss.} 
Going beyond MIMIC-IV, models developed on the proposed dataset could foreshadow the development of more comprehensive models on more specialized datasets such as \cite{ter2021cohort}, for example including molecular data as input and/or prediction targets, to address more fine-grained diagnostic or deterioration prediction targets.
\added{Another promising avenue involves using multiple waveform snapshots over time for deterioration prediction, rather than relying solely on a single observation, thereby capturing dynamic changes in patient conditions more effectively.} 

\added{Finally, our work naturally focuses on quantitative accuracy. However, there are many quality dimensions beyond quantitative accuracy that are important for clinical applications, such as explainability, different aspects of robustness, uncertainty quantification or privacy. As a specific example, it would be interesting to investigate whether specific ECG patterns contribute to the performance gains observed when using full ECG waveforms instead of extracted features. Related methods for explaining ECG-based predictions have been explored in \cite{wagner2023explaining}, and further work could extend these approaches to assess causal versus associational attributions along the lines of \cite{alcaraz2024causalconceptts}. As a second specific example, even though we work with de-identified structured data in this work, it is important to develop models in line with privacy requirements. Here, emerging privacy-preserving techniques, as detailed for example in \cite{feretzakis2024privacy}, will become essential to ensure compliance in clinical decision support systems, especially with the potential future inclusion of free-text data.}

\section{Potential implications}
Clinically, these models enhance decision-making by providing timely and accurate predictions, reducing diagnostic errors, and improving patient care, especially in high-stakes emergency settings. The open-source nature of the curated dataset fosters collaboration and innovation from diverse research communities, allowing for the development of new models and tools that aim to improve benchmark scores and enable more direct comparisons across models. However, questions of algorithmic fairness remain to be investigated in detail in future work.

\subsection{Data and code availability}

\begin{itemize}
\item Project name: MDS-ED
\item Project home page: \cite{Lopez_Alcaraz_Strodthoff_2024}.
\item Project code repository: \added{\url{https://github.com/AI4HealthUOL/MDS-ED}}
\item Operating system(s): Platform independent
\item Programming language: Python
\item Other requirements: Python 3.10.8, PyTorch 1.13.0, and PyTorch Lightening 1.8.0. 
\end{itemize}

\subsection{Author contributions}
JMLA and NS were responsible for the conceptualization of the project. Data curation was performed by JMLA and NS. JMLA and NS implemented the prediction models, with JMLA conducting the formal analysis under NS's supervision. HB provided critical revisions for clinical content. JMLA produced the original draft, and NS and HB contributed to the review and editing. All authors participated in the investigation, methodology, and validation, and they approved the final version for publication. Computing resources were provided through Carl von Ossietzky University of Oldenburg.

\subsection{Generative AI statement}
\added{Generative AI was used for grammar correction but not for text generation in this manuscript.}

\subsection{Competing interests}
The authors have declared no competing interests.

\subsection{Ethics statement}
\added{This study used publicly available and deidentified datasets. No direct patient interaction or intervention was involved. As these datasets are released under established data use agreements and have been ethically approved for secondary research, no additional ethics approval was sought.}

\bibliographystyle{elsarticle-num} 
\bibliography{bibfile}

\begin{thebibliography}{10}
\expandafter\ifx\csname url\endcsname\relax
  \def\url#1{\texttt{#1}}\fi
\expandafter\ifx\csname urlprefix\endcsname\relax\def\urlprefix{URL }\fi
\expandafter\ifx\csname href\endcsname\relax
  \def\href#1#2{#2} \def\path#1{#1}\fi

\bibitem{Rajpurkar2022}
P.~Rajpurkar, E.~Chen, O.~Banerjee, E.~J. Topol, Ai in health and medicine, Nature Medicine 28~(1) (2022) 31–38, doi: \url{10.1038/s41591-021-01614-0}.

\bibitem{johnson2021precision}
K.~B. Johnson, W.-Q. Wei, D.~Weeraratne, M.~E. Frisse, K.~Misulis, K.~Rhee, J.~Zhao, J.~L. Snowdon, Precision medicine, ai, and the future of personalized health care, Clinical and Translational Science 14~(1) (2021) 86--93, doi: \url{10.1111/cts.12884}.

\bibitem{dara2022machine}
S.~Dara, S.~Dhamercherla, S.~S. Jadav, C.~M. Babu, M.~J. Ahsan, Machine learning in drug discovery: a review, Artificial Intelligence Review 55~(3) (2022) 1947--1999, doi: \url{10.1007/s10462-021-10058-4}.

\bibitem{hong2020prediction}
S.~Hong, S.~Lee, J.~Lee, W.~C. Cha, K.~Kim, Prediction of cardiac arrest in the emergency department based on machine learning and sequential characteristics: model development and retrospective clinical validation study, JMIR Medical Informatics 8~(8) (2020) e15932, doi: \url{10.2196/15932}.

\bibitem{harrou2020forecasting}
F.~Harrou, A.~Dairi, F.~Kadri, Y.~Sun, Forecasting emergency department overcrowding: A deep learning framework, Chaos, Solitons \& Fractals 139 (2020) 110247, doi: \url{10.1016/j.chaos.2020.110247}.

\bibitem{covino2020predicting}
M.~Covino, C.~Sandroni, M.~Santoro, L.~Sabia, B.~Simeoni, M.~G. Bocci, V.~Ojetti, M.~Candelli, M.~Antonelli, A.~Gasbarrini, et~al., Predicting intensive care unit admission and death for covid-19 patients in the emergency department using early warning scores, Resuscitation 156 (2020) 84--91, doi: \url{10.1016/j.resuscitation.2020.08.124}.

\bibitem{wang2019machine}
P.~Wang, Y.~Li, C.~K. Reddy, Machine learning for survival analysis: A survey, ACM Computing Surveys (CSUR) 51~(6) (2019) 1--36, doi: \url{10.1145/3214306}.

\bibitem{henna2022interpretable}
S.~Henna, J.~M.~L. Alcaraz, From interpretable filters to predictions of convolutional neural networks with explainable artificial intelligence, ArXiv Preprint arXiv:2207.12958 (2022).

\bibitem{alcaraz2024cardiolablaboratoryvaluesestimation}
J.~M.~L. Alcaraz, N.~Strodthoff, Cardiolab: Laboratory values estimation from electrocardiogram features -- an exploratory study, ArXiv Preprint arXiv:2407.18629 (2024).

\bibitem{taylor2018predicting}
R.~A. Taylor, C.~L. Moore, K.-H. Cheung, C.~Brandt, Predicting urinary tract infections in the emergency department with machine learning, PloS One 13~(3) (2018) e0194085, doi: \url{10.1371/journal.pone.0194085}.

\bibitem{hwang2019deep}
E.~J. Hwang, J.~G. Nam, W.~H. Lim, S.~J. Park, Y.~S. Jeong, J.~H. Kang, E.~K. Hong, T.~M. Kim, J.~M. Goo, S.~Park, et~al., Deep learning for chest radiograph diagnosis in the emergency department, Radiology 293~(3) (2019) 573--580, doi: \url{10.1148/radiol.2019191225}.

\bibitem{strodthoff2024prospects}
N.~Strodthoff, J.~M. Lopez~Alcaraz, W.~Haverkamp, Prospects for artificial intelligence-enhanced electrocardiogram as a unified screening tool for cardiac and non-cardiac conditions: an explorative study in emergency care, European Heart Journal-Digital Health (2024) ztae039Doi: \url{10.1093/ehjdh/ztae039}.

\bibitem{Fleuren2020}
L.~M. Fleuren, T.~L.~T. Klausch, C.~L. Zwager, L.~J. Schoonmade, T.~Guo, L.~F. Roggeveen, E.~L. Swart, A.~R.~J. Girbes, P.~Thoral, A.~Ercole, M.~Hoogendoorn, P.~W.~G. Elbers, \href{http://dx.doi.org/10.1007/s00134-019-05872-y}{Machine learning for the prediction of sepsis: a systematic review and meta-analysis of diagnostic test accuracy}, Intensive Care Medicine 46~(3) (2020) 383–400.
\newblock \href {https://doi.org/10.1007/s00134-019-05872-y} {\path{doi:10.1007/s00134-019-05872-y}}.
\newline\urlprefix\url{http://dx.doi.org/10.1007/s00134-019-05872-y}

\bibitem{sun2024edcopilot}
L.~Sun, A.~Agarwal, A.~Kornblith, B.~Yu, C.~Xiong, \href{https://proceedings.mlr.press/v235/sun24a.html}{{ED}-copilot: Reduce emergency department wait time with language model diagnostic assistance}, in: R.~Salakhutdinov, Z.~Kolter, K.~Heller, A.~Weller, N.~Oliver, J.~Scarlett, F.~Berkenkamp (Eds.), Proceedings of the 41st International Conference on Machine Learning, Vol. 235 of Proceedings of Machine Learning Research, PMLR, 2024, pp. 46942--46956.
\newline\urlprefix\url{https://proceedings.mlr.press/v235/sun24a.html}

\bibitem{sundrani2023predicting}
S.~Sundrani, J.~Chen, B.~T. Jin, Z.~S.~H. Abad, P.~Rajpurkar, D.~Kim, Predicting patient decompensation from continuous physiologic monitoring in the emergency department, NPJ Digital Medicine 6~(1) (2023) 60, doi: \url{10.1038/s41746-023-00803-0}.

\bibitem{wornow2024ehrshot}
M.~Wornow, R.~Thapa, E.~Steinberg, J.~Fries, N.~Shah, Ehrshot: An ehr benchmark for few-shot evaluation of foundation models, Advances in Neural Information Processing Systems 36 (2024).

\bibitem{hager2024evaluation}
P.~Hager, F.~Jungmann, R.~Holland, K.~Bhagat, I.~Hubrecht, M.~Knauer, J.~Vielhauer, M.~Makowski, R.~Braren, G.~Kaissis, et~al., Evaluation and mitigation of the limitations of large language models in clinical decision-making, Nature medicine 30~(9) (2024) 2613--2622.

\bibitem{NEURIPS2023_8f61049e}
E.~Chen, A.~Kansal, J.~Chen, B.~T. Jin, J.~Reisler, D.~E. Kim, P.~Rajpurkar, Multimodal clinical benchmark for emergency care (mc-bec): A comprehensive benchmark for evaluating foundation models in emergency medicine, in: A.~Oh, T.~Naumann, A.~Globerson, K.~Saenko, M.~Hardt, S.~Levine (Eds.), Advances in Neural Information Processing Systems, Vol.~36, Curran Associates, Inc., 2023, pp. 45794--45811.

\bibitem{lee2024multimodal}
S.~A. Lee, S.~Jain, A.~Chen, A.~Biswas, J.~Fang, A.~Rudas, J.~N. Chiang, Multimodal clinical pseudo-notes for emergency department prediction tasks using multiple embedding model for ehr (meme), ArXiv Preprint arXiv:2402.00160 (2024).

\bibitem{MIMICIVECG2023}
B.~Gow, T.~Pollard, L.~A. Nathanson, A.~Johnson, B.~Moody, C.~Fernandes, N.~Greenbaum, J.~W. Waks, P.~Eslami, T.~Carbonati, A.~Chaudhari, E.~Herbst, D.~Moukheiber, S.~Berkowitz, R.~Mark, S.~Horng, Mimic-iv-ecg: Diagnostic electrocardiogram matched subset, doi: \url{10.13026/4nqg-sb35} (2023).
\newblock \href {https://doi.org/10.13026/4NQG-SB35} {\path{doi:10.13026/4NQG-SB35}}.

\bibitem{Johnson2023}
A.~E.~W. Johnson, L.~Bulgarelli, L.~Shen, A.~Gayles, A.~Shammout, S.~Horng, T.~J. Pollard, S.~Hao, B.~Moody, B.~Gow, L.~wei H.~Lehman, L.~A. Celi, R.~G. Mark, {MIMIC}-{IV}, a freely accessible electronic health record dataset, Scientific Data 10~(1), doi: \url{10.1038/s41597-022-01899-x} (Jan. 2023).
\newblock \href {https://doi.org/10.1038/s41597-022-01899-x} {\path{doi:10.1038/s41597-022-01899-x}}.

\bibitem{Lopez_Alcaraz_Strodthoff_2024}
J.~M. Lopez~Alcaraz, N.~Strodthoff, Mimic-iv-ext-mds-ed: Multimodal decision support in the emergency department - a benchmark dataset for diagnoses and deterioration prediction in emergency medicine (version 1.0.0), doi: \url{https://doi.org/10.13026/p90d-vd84} (2024).
\newblock \href {https://doi.org/10.13026/p90d-vd84} {\path{doi:10.13026/p90d-vd84}}.

\bibitem{Mitchell2022}
O.~J.~L. Mitchell, M.~Dewan, H.~A. Wolfe, K.~J. Roberts, S.~Neefe, G.~Lighthall, N.~A. Sands, G.~Weissman, J.~Ginestra, M.~G.~S. Shashaty, W.~D. Schweickert, B.~S. Abella, Defining physiological decompensation: An expert consensus and retrospective outcome validation, Critical Care Explorations 4~(4) (2022) e0677, doi: \url{10.1097/cce.0000000000000677}.
\newblock \href {https://doi.org/10.1097/cce.0000000000000677} {\path{doi:10.1097/cce.0000000000000677}}.

\bibitem{moor2023predicting}
M.~Moor, N.~Bennett, D.~Ple{\v{c}}ko, M.~Horn, B.~Rieck, N.~Meinshausen, P.~B{\"u}hlmann, K.~Borgwardt, Predicting sepsis using deep learning across international sites: a retrospective development and validation study, EClinicalMedicine 62, doi: \url{10.1016/j.eclinm.2023.102124} (2023).

\bibitem{wang2023s4sleep}
T.~Wang, N.~Strodthoff, \href{http://dx.doi.org/10.1016/j.compbiomed.2025.109735}{S4sleep: Elucidating the design space of deep-learning-based sleep stage classification models}, Computers in Biology and Medicine 187 (2025) 109735.
\newblock \href {https://doi.org/10.1016/j.compbiomed.2025.109735} {\path{doi:10.1016/j.compbiomed.2025.109735}}.
\newline\urlprefix\url{http://dx.doi.org/10.1016/j.compbiomed.2025.109735}

\bibitem{mehari2023towards}
T.~Mehari, N.~Strodthoff, Towards quantitative precision for ecg analysis: Leveraging state space models, self-supervision and patient metadata, IEEE Journal of Biomedical and Health InformaticsDoi: \url{10.1109/JBHI.2023.3310989} (2023).

\bibitem{Gu2021EfficientlyML}
A.~Gu, K.~Goel, C.~R{\'e}, Efficiently modeling long sequences with structured state spaces, International Conference on Learning Representations (2021).

\bibitem{loshchilov2018decoupled}
I.~Loshchilov, F.~Hutter, \href{https://openreview.net/forum?id=Bkg6RiCqY7}{Decoupled weight decay regularization}, in: International Conference on Learning Representations, 2019.
\newline\urlprefix\url{https://openreview.net/forum?id=Bkg6RiCqY7}

\bibitem{mcdermott2024closer}
M.~B. McDermott, H.~Zhang, L.~H. Hansen, G.~Angelotti, J.~Gallifant, \href{https://openreview.net/forum?id=S3HvA808gk}{A closer look at {AUROC} and {AUPRC} under class imbalance}, in: The Thirty-eighth Annual Conference on Neural Information Processing Systems, 2024.
\newline\urlprefix\url{https://openreview.net/forum?id=S3HvA808gk}

\bibitem{laRoiTeeuw2024}
H.~M. la~Roi-Teeuw, F.~S. van Royen, A.~de~Hond, A.~Zahra, S.~de~Vries, R.~Bartels, A.~J. Carriero, S.~van Doorn, Z.~S. Dunias, I.~Kant, T.~Leeuwenberg, R.~Peters, L.~Veerhoek, M.~van Smeden, K.~Luijken, Don’t be misled: 3 misconceptions about external validation of clinical prediction models, Journal of Clinical Epidemiology 172 (2024) 111387, doi: \url{10.1016/j.jclinepi.2024.111387}.
\newblock \href {https://doi.org/10.1016/j.jclinepi.2024.111387} {\path{doi:10.1016/j.jclinepi.2024.111387}}.

\bibitem{ter2021cohort}
E.~Ter~Avest, B.~C. van Munster, R.~J. van Wijk, S.~Tent, S.~Ter~Horst, T.~T. Hu, L.~E. van Heijst, F.~S. van~der Veer, F.~E. van Beuningen, J.~C. Ter~Maaten, et~al., Cohort profile of acutelines: A large data/biobank of acute and emergency medicine, BMJ Open 11~(7) (2021) e047349, doi: \url{10.1136/bmjopen-2020-047349}.

\bibitem{wagner2023explaining}
P.~Wagner, T.~Mehari, W.~Haverkamp, N.~Strodthoff, Explaining deep learning for ecg analysis: Building blocks for auditing and knowledge discovery, arXiv preprint arXiv:2305.17043 (2023).

\bibitem{alcaraz2024causalconceptts}
J.~M.~L. Alcaraz, N.~Strodthoff, Causalconceptts: Causal attributions for time series classification using high fidelity diffusion models, ArXiv Preprint arXiv:2405.15871 (2024).

\bibitem{feretzakis2024privacy}
G.~Feretzakis, K.~Papaspyridis, A.~Gkoulalas-Divanis, V.~S. Verykios, Privacy-preserving techniques in generative ai and large language models: A narrative review, Information 15~(11) (2024) 697.

\end{thebibliography}

\clearpage

\appendix

\section{Effect of the binary NaN mask on model performance}\label{app:mask_ablation}

\added{For the ECG waveforms + clinical routine data setting (deep learning), our main results achieved 0.8256 (0.8222, 0.8288) for the diagnoses model and 0.9115 (0.8991, 0.9222) for the deterioration model when using a binary NaN mask column. In this section, we present additional experiments as a small ablation study to demonstrate that omitting the binary NaN mask column leads to inferior performance. Specifically, without the mask, the models achieved 0.8150 (0.8113, 0.8185) and 0.8986 (0.8864, 0.9094), respectively.}

\section{Advantages of multimodal integration in clinical settings}\label{app:advantages_multimodal}

\added{In Table~\ref{tab:multimodalvs1}, we present a more finegrained performance comparison, which disentangles the overall performance according to ICD chapters in the diagnostic case and according to three categories, see Table~\ref{tab:multimodalvs1} for the assignment of individual deterioration labels to these categories. In Table~\ref{tab:multimodalvs2}, we show a detailed breakdown of the model performance within the deterioration case. We focus on highlighting the added value of including full waveforms (ECG W. \& CR) as opposed to waveform features (ECG F. \& CR). We indicate the relative improvement of the former over the latter setting in the final column in both tables. Most notably, including full waveforms yields performance improvements for every single entry in both tables.}

\begin{table}[ht!]
\centering
\scriptsize
\caption{\added{Detailed breakdown of model performance for different ICD-10 categories and clinical deterioration outcomes. Unimodal settings are CR: features derived from clinical routine data, ECG F.: ECG features, and ECG W.: ECG waveforms. Multimodal settings are ECG F. \& CR: ECG features and clinical routine data, as well as ECG W. \& CR: ECG waveforms and clinical routine data. Finally, Relative Improvement (\%): relative improvement of ECG W. \& CR against ECG F. \& CR.}}
\label{tab:multimodalvs1}
\begin{tabular}{lllllll}
\toprule
& \textbf{CR} & \textbf{ECG F.} & \textbf{ECG W.} & \textbf{ECG F. \& CR} & \textbf{ECG W. \& CR} & \textbf{Relative Improvement (\%)} \\ 
\hline
\midrule
IX: Circulatory         & 0.7645 & 0.7153 & 0.8050  & 0.8050 & 0.8761 & 8.83  \\
II: Neoplasms          & 0.8109 & 0.5886 & 0.8101 & 0.8101 & 0.8672 & 7.05  \\
XIX: Injury, Poisoning & 0.7425 & 0.6262 & 0.7712 & 0.7712 & 0.8479 & 9.95  \\
I: Infections          & 0.7485 & 0.6258 & 0.7769 & 0.7769 & 0.8346 & 7.43  \\
IV: Endocrine         & 0.7723 & 0.6401 & 0.7920 & 0.7920 & 0.8336 & 5.25  \\
III: Blood            & 0.8048 & 0.5992 & 0.8143 & 0.8143 & 0.8335 & 2.36  \\
XI: Digestive         & 0.7510 & 0.5902 & 0.7604 & 0.7604 & 0.8270 & 8.76  \\
XIV: Genitourinary    & 0.7743 & 0.6053 & 0.7766 & 0.7766 & 0.8253 & 6.27  \\
X: Respiratory        & 0.7300 & 0.6581 & 0.7643 & 0.7643 & 0.8226 & 7.63  \\
V: Mental            & 0.7398 & 0.6456 & 0.7619 & 0.7619 & 0.8111 & 6.46  \\
XII: Skin            & 0.6951 & 0.6284 & 0.7153 & 0.7153 & 0.8087 & 13.06 \\
VI: Nervous          & 0.7088 & 0.5870 & 0.7183 & 0.7183 & 0.7955 & 10.75 \\
XVIII: Symptoms      & 0.6983 & 0.6090 & 0.7168 & 0.7168 & 0.7813 & 9.00  \\
XIII: Musculoskeletal & 0.6786 & 0.5939 & 0.6930 & 0.6930 & 0.7405 & 6.85  \\
\midrule
Clinical Deterioration & 0.7729 & 0.7135 & 0.7515 & 0.8294 & 0.9070 & 9.36  \\
ICU Admission          & 0.8738 & 0.7377 & 0.7474 & 0.8865 & 0.9063 & 2.23  \\
Mortality             & 0.8792 & 0.7579 & 0.8160 & 0.8920 & 0.9168 & 2.78  \\
\bottomrule
\end{tabular}
\end{table}

\begin{table}[ht!]
\centering
\scriptsize
\caption{\added{Detailed breakdown of model performance category for all investigated sets of input features in the deterioration setting. Unimodal settings are CR: features derived from clinical routine data, ECG F.: ECG features, and ECG W.: ECG waveforms. Multimodal settings are ECG F. \& CR: ECG features and clinical routine data, as well as ECG W. \& CR: ECG waveforms and clinical routine data. Finally, Relative Improvement (\%): relative improvement of ECG W. \& CR against ECG F. \& CR. We summarize the first six entries as `Clinical Deterioration', the following two as `ICU Admission' and the remaining seven as `Mortality'.}}
\label{tab:multimodalvs2}
\begin{tabular}{lllllll}
\toprule
& \textbf{CR} & \textbf{ECG F.} & \textbf{ECG W.} & \textbf{ECG F. \& CR} & \textbf{ECG W. \& CR} & \textbf{Relative Improvement (\%)} \\ 
\hline
\midrule
Severe hypoxemia        & 0.4834 & 0.5311 & 0.7030 & 0.5055 & 0.6980 & 38.08 \\
ECMO                    & 0.7873 & 0.6126 & 0.7768 & 0.8991 & 0.9355 & 4.05  \\
Vasopressors            & 0.8241 & 0.7759 & 0.7672 & 0.8551 & 0.9239 & 8.05  \\
Inotropes               & 0.6550 & 0.7900 & 0.6184 & 0.8064 & 0.9400 & 16.57 \\
Mechanical ventilation  & 0.9373 & 0.7262 & 0.7673 & 0.9418 & 0.9590 & 1.83  \\
IHCA                    & 0.9502 & 0.8454 & 0.8760 & 0.9682 & 0.9859 & 1.83  \\
\midrule
ICU 24 hours           & 0.8809 & 0.7390 & 0.7443 & 0.8928 & 0.9147 & 2.45  \\
ICU overall            & 0.8666 & 0.7363 & 0.7506 & 0.8801 & 0.8979 & 2.02  \\
\midrule
In-hospital            & 0.8371 & 0.8753 & 0.8715 & 0.8736 & 0.9423 & 7.86  \\
24 hours               & 0.9304 & 0.7267 & 0.8409 & 0.9304 & 0.9600 & 3.18  \\
7 days                 & 0.9176 & 0.7266 & 0.8176 & 0.9286 & 0.9429 & 1.54  \\
28 days                & 0.8858 & 0.7329 & 0.7859 & 0.8946 & 0.9115 & 1.89  \\
90 days                & 0.8651 & 0.7590 & 0.7926 & 0.8792 & 0.8952 & 1.82  \\
180 days               & 0.8605 & 0.7464 & 0.8036 & 0.8727 & 0.8894 & 1.91  \\
365 days               & 0.8579 & 0.7387 & 0.7999 & 0.8646 & 0.8768 & 1.41  \\
\bottomrule
\end{tabular}
\end{table}

\section{Dataset details}\label{app:dataset}

\begin{table}[ht!]
\centering
\scriptsize

\caption{Definitions of labels for the diagnoses and deterioration models, including 1428 discharge diagnoses based on ICD10-CM codes in the diagnoses model, and 15 labels covering clinical outcomes such as mortalities, ICU admission, vasopressors, inotropes, mechanical ventilation, in-hospital cardiac arrest (IHCA), extracorporeal membrane oxygenation (ECMO), and severe hypoxemia in the deterioration model.}
\label{tab:tasks}
\begin{tabular}{p{2cm} c p{4cm}}
\toprule
\textbf{Name} & \textbf{Labels} & \textbf{Details} \\  
\hline
\midrule
\multicolumn{3}{l}{\textbf{Diagnoses model (1428 labels, based on \cite{strodthoff2024prospects})}} \\ 
\midrule
Diagnoses & 1428 & Discharge diagnoses ICD10-CM codes \\
\midrule
\multicolumn{3}{l}{\textbf{Deterioration model (15 labels, based on \cite{Mitchell2022})}} \\ 
\midrule
Mortalities & 6 & 1, 7, 30, 90, 180, 365 days and in-stay \\
ICU Admission & 2 & 24 hours and in-stay \\
Vasopressors & 1 & Within 24 hours: epinephrine, norepinephrine, vasopressin, dobutamine, dopamine, or phenylephrine \\
Inotropes & 1 & Within 24 hours: epinephrine, dobutamine, or dopamine \\
Mechanical Ventilation & 1 & Same day or next day ICD9/ICD10 codes: 9670, 9671, 9672, 5A1935Z, 5A1945Z, and 5A1955Z \\
In-Hospital Cardiac Arrest (IHCA) & 1 & Same day or next day ICD9/ICD10 codes: I469, 4275, I462, V1253, I468 \\
Extracorporeal Membrane Oxygenation (ECMO) & 1 & Same day or next day ICD9/ICD10 codes: 3961, 3965, 3966, 5A1221Z, 5A1522G, 5A1522H, 5A15223, 5A1522F, 5A15A2F, 5A15A2G, and 5A15A2H \\
Severe Hypoxemia & 1 & Oxygen saturation $\leq$ 85\% within 24 hours \\
\bottomrule
\end{tabular}
\end{table}

Table \ref{tab:tasks} contains the considered tasks by description, number of labels, and additional details such as target criteria definition. Apart from diagnoses, we use the rest of the tasks defined as patient deterioration based on previous work from the literature that defines physiological deterioration \cite{Mitchell2022}. Although we consider diverse scores such as SOFA significant, calculating these markers requires a more detailed definition of downstream task prediction, which goes beyond the scope of maintaining homogeneity across tasks in this study. The curation of the dataset starts from MIMIC-IV-ECG-ICD \cite{strodthoff2024prospects}, from which we obtained the ICD10 diagnoses codes for the complete MIMIC-IV-ECG \cite{MIMICIVECG2023} dataset, therefore, based on ECG records with available ED stay\_id, we select the ones that happened between the patient in time and the 1.5 hours window. Mortalities at finite horizons were computed based on the date of death from \cite{MIMICIVECG2023}, and for the computation of mortality per stay, we include discharge date times from the \texttt{admissions} table in \cite{Johnson2023} MIMIC-IV to account for final discharge, all of the mortalities dates were previously validated from the source with a follow-up up to 1 year after the patient's last recorded discharge. ICU admission in 24 hours and per stay was computed based on the \texttt{ICU stays} table from \cite{Johnson2023}. Mechanical ventilation and ECMO were computed given the \texttt{procedures} table from MIMIC-IV-ED given the patient admission and procedure date. Cardiac arrest was computed from the \texttt{diagnosis} tables from MIMIC-IV-ED and MIMIC-IV, where we first validated if the patient was discharged with the diagnosis code within the ED or hospital otherwise. We clarify that although mechanical ventilation, ECMO, and cardiac arrest do not provide an exact time of events, we carefully approximate the times with dates given the same day or the day after. Vasopressors and inotropes were computed from the \texttt{pyxis} table from MIMIC-IV-ED given the exact time of the applied medication. Finally, severe hypoxemia was computed from the vital signs table from MIMIC-IV-ED. In the deterioration setting, to not discard samples where the events happened before the 1.5 hours for a single task, e.g. severe hypoxemia within 1.5 hours, we include a special token for the task in the label space which we mask during training and evaluation in order not to negatively affect training and evaluation for other labels due to exclusion of entire patients, see \ref{app:models} for more details.


\begin{table}[ht!]
\centering
\scriptsize
\caption{Summary demographic characteristics across samples including gender distribution, age distribution (median with standard deviation and quartiles), and ethnicity distribution in the study population, presented as percentages.}
\label{tab:demographics}
\begin{tabular}{ll}
\toprule
\textbf{Name} & \textbf{Values} \\ 
\midrule
\multicolumn{2}{l}{\textbf{Gender (\%)}} \\ 
\midrule
Female & 68289 (52.9) \\ 
Male & 60806 (47.1) \\ 
\midrule
\multicolumn{2}{l}{\textbf{Age (\%)}} \\ 
\midrule
Median years (SD) & 64 (28) \\ 
Quantile 1: 18-49 & 32634 (25.27) \\ 
Quantile 2: 50-64 & 34376 (26.62) \\ 
Quantile 3: 65-77 & 31821 (24.64) \\ 
Quantile 4: 78-101 & 30264 (23.44) \\ 
\midrule
\multicolumn{2}{l}{\textbf{Ethnicity (\%)}} \\ 
\midrule
White & 78310 (60.66) \\ 
Black & 27680 (21.44) \\ 
Hispanic & 9230 (7.14) \\ 
Asian & 4700 (3.64) \\ 
Other & 9175 (7.10) \\ 
\bottomrule
\end{tabular}
\end{table}

Table \ref{tab:demographics} provides detailed demographic information on the study population, including gender, age, and ethnicity. The gender distribution shows a slightly higher proportion of females (52.9\%) compared to males (47.1\%). The age distribution is divided into four quantiles, with a median age of 64 years and a standard deviation of 28 years. Regarding ethnicity, the majority of the population is White (60.66\%), followed by Black (21.44\%), Hispanic (7.14\%), Asian (3.64\%), and individuals of other ethnicities (7.10\%). This diverse demographic distribution ensures a comprehensive evaluation of the model across various patient groups. These variables were extracted from the \texttt{ed stays} table from MIMIC-IV-ED.


\begin{table*}[!ht]
\centering
\scriptsize
\caption{Summary of statistics across samples for biometric measurements, vital parameters, and triage acuity levels, including minimum, maximum, median, interquartile range (IQR), and number of records for each variable.}
\label{tab:biometrics_vital_parameters}
\begin{tabular}{lllllll}
\toprule
\textbf{Name} & \textbf{Unit} & \textbf{Minimum} & \textbf{Maximum} & \textbf{Median} & \textbf{IQR} & \textbf{Records} \\ 
\midrule
\multicolumn{7}{l}{\textbf{Biometrics}} \\ 
\midrule
BMI & kg/m$^2$ & 0 & 99.2 & 27.7 & 8.70 & 71234 \\
Weight & kg & 20.41 & 367.99 & 77.11 & 27.56 & 74026 \\
Height & cm & 64.99 & 387.09 & 167.64 & 15.23 & 42084 \\
\midrule
\multicolumn{7}{l}{\textbf{Vital parameters}} \\
\midrule
Temperature & $^\circ$C & 13.3 & 44.72 & 36.6 & 0.55 & 77179 \\
Heart rate & bpm & 0.0 & 570.0 & 84.0 & 30.0 & 153249 \\
Respiratory rate & bpm & 0.0 & 169.0 & 18.0 & 4.0 & 150525 \\
Oxygen saturation & \% & 0.0 & 100.0 & 99.0 & 3.0 & 146140 \\
SBP & mmHg & 0.0 & 274.0 & 130.0 & 35.0 & 151346 \\
DBP & mmHg & 0.0 & 486.0 & 73.0 & 23.0 & 151254 \\
\midrule
\multicolumn{7}{l}{\textbf{Triage}} \\
\midrule
Acuity & level & 1 & 5 & 2 & 1 & 126071 \\
\bottomrule
\end{tabular}
\end{table*}

Table \ref{tab:biometrics_vital_parameters} provides details on the biometrics, vital parameters, and triage values considered in this work, including the name, unit, minimum value, maximum value, median value, and the number of records. Dataset preprocessing includes weight and height conversion from pounds and inches to kilograms and centimeters. During the data cleaning process, we removed outliers and introduced missing values at various thresholds to ensure model computational convergence. The specific thresholds for outliers were as follows: temperature below and above 50 and 150 Fahrenheit degrees respectively. Heart rate above 700, respiration rate above 300, oxygen saturation below 0 and above 100, diastolic and systolic blood pressures above 500. BMI above 100, weight below 20 or above 400, and height below 60 and above 400. The temperature unit was converted from Fahrenheit to Celsius, whereas weight from pounds to kilograms, and height from inches to centimeters. Biometrics features were extracted from the online medical records table from MIMIC-IV from which we match the closest value of a patient stay within 30 days of admission, otherwise missing. Vital parameters and acuity were extracted from the \texttt{vital signs} and \texttt{triage} tables of MIMIC-IV-ED.


\begin{table*}[!ht]
\centering
\scriptsize
\caption{Summary of laboratory values across samples considered in this study, including the name, unit, minimum value, maximum value, median value, fluid type, and the number of records for each parameter.}
\label{tab:lab_values}
\begin{tabular}{llllllll}
\toprule
\textbf{Name} & \textbf{Unit} & \textbf{Minimum} & \textbf{Maximum} & \textbf{Median} & \textbf{IQR} & \textbf{Fluid} & \textbf{Records} \\ 
\midrule
Abs. Basophil Count & K/uL & 0.0 & 9.2 & 0.04 & 0.03 & Blood & 36837 \\
Abs. Eosinophil Count & K/uL & 0.0 & 15.33 & 0.09 & 0.15 & Blood & 36838 \\
Abs. Lymphocyte Count & K/uL & 0.0 & 88.08 & 1.47 & 1.12 & Blood & 36909 \\
Alanine Aminotransf. & IU/L & 1.0 & 1999.0 & 22.0 & 22.0 & Blood & 29999 \\
Albumin & g/dL & 1.0 & 6.7 & 4.0 & 0.79 & Blood & 28324 \\
Alkaline Phosphatase & IU/L & 5.0 & 1985.0 & 85.0 & 51.0 & Blood & 29923 \\
Asparate Aminotransf. & IU/L & 1.0 & 1996.0 & 29.0 & 29.0 & Blood & 29995 \\
Bands & \% & 0.0 & 61.0 & 0.0 & 3.0 & Blood & 6057 \\
Base Excess & mEq/L & -413.0 & 29.0 & 0.0 & 6.0 & Blood & 8011 \\
Basophils & \% & 0.0 & 35.0 & 0.4 & 0.39 & Blood & 79386 \\
Bicarbonate & mEq/L & 2.0 & 50.0 & 24.0 & 5.0 & Blood & 79771 \\
Bilirubin, Direct & mg/dL & 0.0 & 45.5 & 1.0 & 2.5 & Blood & 1174 \\
Bilirubin, Total & mg/dL & 0.0 & 55.8 & 0.5 & 0.5 & Blood & 30098 \\
C-Reactive Protein & mg/L & 0.1 & 299.3 & 17.95 & 75.14 & Blood & 1108 \\
Calcium, Total & mg/dL & 0.0 & 79.2 & 9.2 & 0.79 & Blood & 30262 \\
Carboxyhemoglobin & \% & 0.0 & 18.0 & 2.0 & 2.0 & Blood & 1078 \\
Chloride & mEq/L & 58.0 & 140.0 & 102.0 & 6.0 & Blood & 79855 \\
Creatine Kinase (CK) & IU/L & 8.0 & 1998.0 & 115.0 & 158.0 & Blood & 9588 \\
CK, MB Isoenzyme & ng/mL & 1.0 & 497.0 & 3.0 & 4.0 & Blood & 9272 \\
Creatinine & mg/dL & 0.0 & 43.0 & 1.0 & 0.5 & Blood & 82850 \\
Eosinophils & \% & 0.0 & 75.0 & 1.3 & 2.1 & Blood & 79387 \\
Fibrinogen, Functional & mg/dL & 35.0 & 1348.0 & 298.5 & 161.5 & Blood & 2280 \\
Free Calcium & mmol/L & 0.48 & 3.31 & 1.08 & 0.12 & Blood & 1472 \\
Glucose & mg/dL & 3.8 & 1684.0 & 113.0 & 50.0 & Blood & 85022 \\
Hematocrit & \% & 6.9 & 66.1 & 38.3 & 7.89 & Blood & 82490 \\
Hemoglobin & g/dL & 0.0 & 21.8 & 12.6 & 3.0 & Blood & 84890 \\
INR(PT) & nan & 0.75 & 24.0 & 1.1 & 0.30 & Blood & 44656 \\
Lactate & mmol/L & 0.3 & 25.2 & 1.8 & 1.20 & Blood & 32666 \\
Lymphocytes & \% & 0.0 & 100.0 & 19.7 & 17.1 & Blood & 79387 \\
Magnesium & mg/dL & 0.2 & 45.0 & 2.0 & 0.40 & Blood & 30472 \\
Neutrophils & \% & 0.0 & 99.0 & 69.9 & 19.59 & Blood & 79387 \\
Oxygen Saturation & \% & 0.2 & 901.0 & 80.0 & 32.0 & Blood & 4838 \\
PT & sec & 8.2 & 150.0 & 12.0 & 3.5 & Blood & 44656 \\
PTT & sec & 16.6 & 150.0 & 30.6 & 6.99 & Blood & 44497 \\
Phosphate & mg/dL & 0.2 & 23.8 & 3.4 & 1.1 & Blood & 29888 \\
Platelet Count & K/uL & 5.0 & 1898.0 & 227.0 & 105.0 & Blood & 82260 \\
Potassium & mEq/L & 1.3 & 10.7 & 4.2 & 0.80 & Blood & 79857 \\
RDW & \% & 10.6 & 33.8 & 13.8 & 2.19 & Blood & 82266 \\
Red Blood Cells & m/uL & 0.65 & 8.06 & 4.25 & 0.96 & Blood & 82266 \\
Sodium & mEq/L & 88.0 & 179.0 & 139.0 & 5.0 & Blood & 79865 \\
Troponin T & ng/mL & 0.0 & 17.36 & 0.06 & 0.13 & Blood & 41806 \\
Urea Nitrogen & mg/dL & 1.0 & 263.0 & 18.0 & 13.0 & Blood & 82522 \\
White Blood Cells & K/uL & 0.0 & 632.1 & 8.1 & 4.60 & Blood & 82367 \\
pCO2 & mm Hg & 8.0 & 189.0 & 44.0 & 17.0 & Blood & 8008 \\
pH & units & 5.0 & 9.0 & 6.0 & 1.0 & Urine & 20968 \\
\bottomrule
\end{tabular}
\end{table*}

Table \ref{tab:lab_values} provides details on the laboratory values considered in this work, including the name, unit, minimum value, maximum value, median value, fluid type, and the number of records. During the data cleaning process, we removed outliers and introduced missing values at various thresholds to ensure model computational convergence. The specific thresholds for outliers were as follows: absolute basophil count above 20, absolute eosinophil count above 20, absolute lymphocyte count above 100, alanine aminotransferase above 2000, alkaline phosphatase above 2000, aspartate aminotransferase above 2000, creatine kinase above 2000, glucose above 2000, lactate above 2000, and platelet count above 2000. Laboratory values were extracted from the \texttt{laboratory events} table from MIMIC-IV, where for each record we matched all the patient records of the desired tests available between patient record in-time and the 1.5 hours window.

\section{Models}\label{app:models}

\begin{table}[!ht]
    \centering
    \scriptsize
    \caption{S4 model hyperparameters}
    \label{tab:s4-hyperparameters}
    \begin{tabular}{ll}
        \toprule
        \textbf{Hyperparameter} & \textbf{Value}    \\ 
        \midrule
        Block of layers & 4 \\
        S4 model copies & 512 \\
        S4 state size & 8 \\ 
        Optimizer & Adam \\ 
        Learning rate & 0.001 \\ 
        Weight decay & 0.001 \\ 
        Learning rate schedule & constant \\ 
        Batch size & 64 \\ 
        Epochs & 20 \\ 
        \bottomrule
    \end{tabular}
\end{table}

Table \ref{tab:s4-hyperparameters} outlines the hyperparameters employed in the S4 model. The architecture consists of four blocks of layers, with each block containing 512 copies of the S4 model. The state size within the S4 model is set to 8. For optimization, the AdamW optimizer is utilized with a learning rate and weight decay both set to 0.001. The learning rate schedule is maintained constant throughout training. A batch size of 64 samples is used for each training iteration, spanning a total of 20 epochs. The training objective is to minimize the binary cross-entropy loss. During training, we apply a model selection strategy on the best performance (AUROC) on the validation set which usually converges earlier than the final number of epochs. As previously described in \ref{app:dataset}, with the goal of not discarding samples whose any of the deterioration targets happens before the 1.5-hour window or the target event cannot be clearly defined, we introduce a special token in the label space for which we account during and training and evaluation to be discarded. As of XGboost models, we fit these models with the default hyperparameter values in a binary setting for the later AUROC aggregation across settings, models which in principle do not intrinsically allow the inclusion of samples during training with the special token to be discarded.

\end{document}